\documentclass{article}

\usepackage{microtype}
\usepackage{graphicx}
\usepackage{booktabs} %

\usepackage{hyperref}

\usepackage{amsmath,amsthm,amssymb}
\usepackage{subcaption}
\usepackage{multirow}
\usepackage{verbatim}

\usepackage{array}
\newcolumntype{L}[1]{>{\raggedright\let\newline\\\arraybackslash\hspace{0pt}}m{#1}}
\newcolumntype{C}[1]{>{\centering\let\\}m{#1}}

\newcolumntype{R}[1]{>{\raggedleft\let\newline\\\arraybackslash\hspace{0pt}}m{#1}}

\usepackage[algo2e, linesnumbered]{algorithm2e}
\usepackage{amsmath,amsthm,amssymb}
\usepackage{subcaption}
\usepackage{multirow}
\usepackage{verbatim}
\usepackage[accepted]{icml2019}
\usepackage{breqn}

\icmltitlerunning{Robust Learning under Label Noise with Iterative Noise-Filtering}

\begin{document}

\twocolumn[
\icmltitle{Robust Learning under Label Noise with Iterative Noise-Filtering}

\icmlsetsymbol{equal}{*}

\begin{icmlauthorlist}
\icmlauthor{Duc Tam Nguyen}{Bo,Freiburg}
\icmlauthor{Thi-Phuong-Nhung Ngo}{BCAI}
\icmlauthor{Zhongyu Lou}{Bo}
\icmlauthor{Michael Klar}{Bo}
\icmlauthor{Laura Beggel}{BCAI}
\icmlauthor{Thomas Brox}{Freiburg}
\end{icmlauthorlist}

\icmlaffiliation{Bo}{Corporate Research, Robert Bosch GmbH, Renningen, Germany}
\icmlaffiliation{BCAI}{Bosch Center for Artificial Intelligence, Germany}
\icmlaffiliation{Freiburg}{Computer Vision Group, University of Freiburg, Freiburg, Germany }

\icmlcorrespondingauthor{Duc Tam Nguyen}{Nguyen@cs.uni-freiburg.de}
\icmlcorrespondingauthor{Zhongyu Lou}{Zhongyu.Lou@de.bosch.com}
\icmlcorrespondingauthor{Laura Beggel}{Laura.Beggel@de.bosch.com}
\icmlcorrespondingauthor{Nhung Ngo}{ThiPhuongNhung.Ngo@de.bosch.com}
\icmlcorrespondingauthor{Michael Klar}{Michael.Klar2@de.bosch.com}
\icmlcorrespondingauthor{Thomas Brox}{Brox@cs.uni-freiburg.de}

\icmlkeywords{Machine Learning}

\vskip 0.3in
]

\printAffiliationsAndNotice{}  %

\begin{abstract}
	We consider the problem of training a model under the presence of label noise. Current approaches identify samples with potentially incorrect labels and reduce their influence on the learning process by either assigning lower weights to them or completely removing them from the training set. In the first case the model however still learns from noisy labels; in the latter approach, good training data can be lost. In this paper, we propose an iterative semi-supervised mechanism for robust learning which excludes noisy labels but is still able to learn from the corresponding samples. To this end, we add an unsupervised loss term that also serves as a regularizer against the remaining label noise. We evaluate our approach on common classification tasks with different noise ratios. Our robust models outperform the state-of-the-art methods by a large margin. Especially for very large noise ratios, we achieve up to $20$\% absolute improvement compared to the previous best model. 
\end{abstract}

\section{Introduction}

\begin{figure}[h]
	\begin{center}
		\includegraphics[width=.9\linewidth]{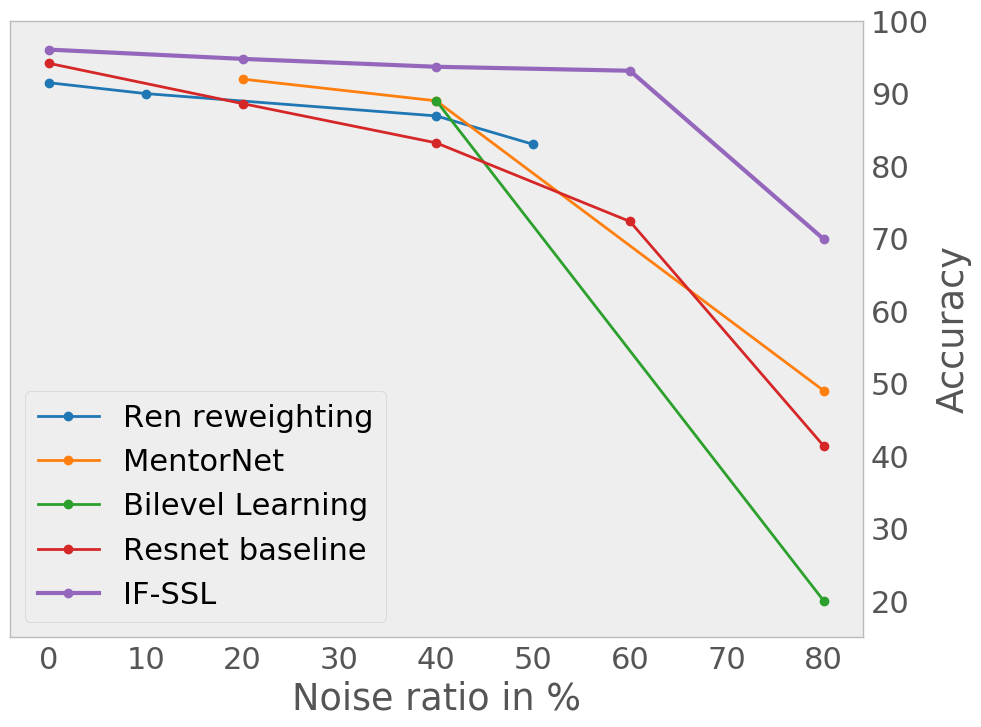}
	\end{center}
	\caption{Noisy CIFAR-10: Model performance measured in accuracy for learning under different label noise ratios. Our framework IF-SSL retains highly robust classification and is only impaired in case of extreme noise such as 80\%. Contrary, the performance of the basic baseline without iterative-filtering and semi-supervised learning rapidly decreases.}
	\label{fig:cifar-10:acc}
\end{figure}
In many supervised learning applications, a clean labeled dataset is the key to success. However, in real-world scenarios, label noise inevitably originates from different sources such as inconsistent labelers or the difficulty of the labeling task itself. 
In many classification tasks for example samples that cannot be squeezed into a strict categorical scheme will lead to inconsistent labels.

With traditional supervised learning, the present label noise decreases the performance of classification models since they tend to over-fit to the samples with noisy labels. This results in lower accuracy and inferior generalization properties. To avoid the negative influence of noisy labels, a common approach is to use sample-dependent loss weights as learning regularizers~\cite{jiang_mentornet:_2017-3,ren_learning_2018-3}. However, the performance of these mechanisms strongly depends on the respective hyperparameters that are difficult to set. 

Typically the loss weights $\mathbf{w}$ are restricted to $\mathbf{w} \in [0,1]$ by design to resemble a probability of a noisy label given a sample. In a supervised learning framework, however, even with tiny (e.g., $0.01$) loss weights, the model could still receive a strong learning signal from noisy samples (as, e.g., in~\cite{ren_learning_2018-3}). A perfect case is assigning weights $\mathbf{w} = 0$  to samples with noisy labels which, however, implies ignoring those samples and results in a smaller training dataset.

In this paper, instead of training in a supervised framework, we 
learn from the samples with noisy labels in an unsupervised way. Since the input data are not noisy but only the labels, semi-supervised learning can still exploit the raw data samples. By keeping those samples rather than removing them from training our proposed method can be more strict when it comes to removing potentially noisy labels.

In more detail, we propose a learning scheme consisting of (1) iterative filtering of noisy labels and (2) semi-supervised learning to regularize the problem in the vicinity of noisy samples. Fig.~\ref{fig:IF-SSL-overview} shows a simplified overview of the concept. We refer to the proposed training procedure as \emph{Iterative Filtering with Semi-Supervised Learning} (IF-SSL). To the best of our knowledge, we propose the first approach that only removes the noisy labels instead of the complete data samples using filtering.  Our approach requires no new dedicated mechanism for robust learning and utilizes only existing standard components for learning.

The proposed algorithm was evaluated on classification tasks for CIFAR-10 \& CIFAR-100 with a varying label noise ratio from 0\% to 80\%.  We show results both for a clean validation set and a noisy one. In both cases, we show that using the filtered data as unlabeled samples significantly outperforms complete removal of the data. As a consequence, the proposed model consistently outperforms state of the art at all levels of label noise; see Fig.~\ref{fig:cifar-10:acc}. Despite the simplicity of the training pipeline, our approach shows robust performance even in case of high noise ratios. 
The source code will be made available together with the published paper.

\section{Robust Learning with Iterative noise-filtering }

\subsection{Overview}
\begin{figure}[t]
	\begin{center}
		\includegraphics[width=1\linewidth]{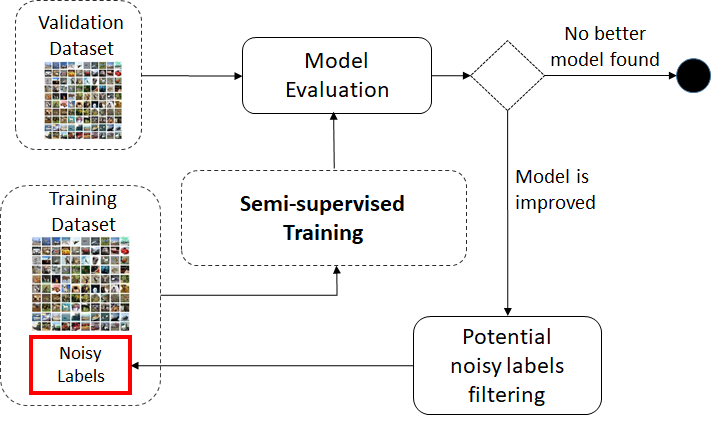}
	\end{center}
	\caption{Overview of the proposed iterative noise-filtering with semi-supervised learning. In each iteration we filter only potentially noisy \emph{labels} and keep using all images for training. This filtering process stops when no further iterative filtering can improve the model's performance on the validation set.}
	\label{fig:IF-SSL-overview}
\end{figure}

Fig.~\ref{fig:IF-SSL-overview} shows an overview of our proposed approach. In the beginning, we assume that the labels of the training set are noisy (up to a certain noise ratio). We use a small validation set to measure the improvement in model performance.
In each iteration, we first apply semi-supervised training until we find the best model w.r.t. the performance on the validation set (e.g., by early-stopping). In the next step, we use the moving-average-prediction results of the best model to filter out potentially noisy labels based on the strategy defined in Section~\ref{ss:general scheme}. In the next iteration, we again use all data and the new filtered label set as input for the model training. The iterative training procedure stops when no better model can be found. Our filtering pipeline only requires a standard component of training deep learning models.

To provide a powerful regularizer against label noise, the semi-supervised model treats all data points as additional unlabeled samples. Concretely, in the first iteration, the model learns from supervised and unsupervised learning objectives on the complete dataset. Subsequently, the unsupervised learning objective continuously derives learning signals from all data points while the supervised learning objective is computed only on a filtered set of labeled samples. Over these iterations, the label noise in the training set is expected to reduce.

In the following, we give more details about the combination of this training and filtering procedure with existing techniques from semi-supervised learning.

\subsection{Iterative Filtering}
\label{ss:general scheme}

Let us start with an initial noisy training dataset $\mathcal{D}_{0}$ and and the validation set $\mathcal{D}_{val}$. Assume each example might have one of the following labels $\{l_1, \ldots, l_m\} \cup \{l_{\emptyset}\}$ where $l_{\emptyset}$ denotes the unlabeled/noisy case.
By $M$ we denote the model which in the current training epoch maps each example $x_i$ to a set $\{l_1:c_1, \ldots, l_m:c_m\}$ where $c_i \geq 0$ is the score and $\Sigma_{i}^{m} c_i = 1$. Let $acc(M, {\mathcal{D}}_{val})$ be the accuracy of $M$ over the validation set ${\mathcal{D}}_{val}$. 

Let $train\_and\_valid({\mathcal D}_{i}, {\mathcal D}_{val})$ denote a training procedure which will be explained in detail in Section~\ref{s:unsupervised}. 

\begin{algorithm}[t]
	\SetAlgoLined
	\SetKwInput{KOutput}{Output}
	\SetKwInput{KInput}{Input}
	\KInput{${\mathcal D}_{0}, {\mathcal D}_{val}$, $n\_max\_iterations$}
	\KOutput{$M_{best}$}
	
	Initialize $M_{best}: = train\_and\_valid({\mathcal D}_{0}, {\mathcal D}_{val})$ 
	
	$i: = 1$\;
	\While{true}{
		$M_i: = train\_and\_valid({\mathcal D}_{i}, {\mathcal D}_{val})$
		
		\If{$acc(M_{best}, {\mathcal{D}}_{val}) \geq acc(M_{i}, {\mathcal{D}}_{val})$ or $i > n\_max\_iterations$}{
			return  $M_{best}$\;
		}
		
		$M_{best} := M_i$\;
		
		${\mathcal D}_{filter} := {\mathcal D}_0$\;
		
		\For{$(x, l_j)$ in ${\mathcal D}_{filter}$}{
			$\{l_1:c_1, \ldots, l_m:c_m\}$: = $M_i(x)$\;
			
			\If{$c_j \neq max(c_i, i = 1,m)$ }{
				Replace $l_j$ by $l_{\emptyset}$ in ${\mathcal D}_{filter}$\;
			}
		}
		
		$i := i+1$\;
		
		${\mathcal D}_i: = {\mathcal D}_{filter}$\;
	}
	\caption{Iterative noisy labels filtering}
	\label{alg:filtering}
\end{algorithm}

Using these notations, the label filtering algorithm is given in Algorithm~\ref{alg:filtering}. 

The label filtering is performed on the \emph{original} label set from iteration $0$. In this way, clean labels erroneously removed in an earlier iteration (e.g., labels of hard to classify samples) can be used for the model training again. This is a major difference to typical iterative filtering approaches where the filtering at iteration $i$ is restricted to training samples from the respective iteration only.

We apply a variant of easy sample mining and filter out training samples based on the model's agreement with the provided label. That means the labels are only used for supervised training if in the current epoch the model predicts the respective label to be the correct class with the highest likelihood. This is reflected in Algorithm~\ref{alg:filtering} line 12 to line 14.  

\begin{figure}
	\centering
	\includegraphics[width=.8\linewidth]{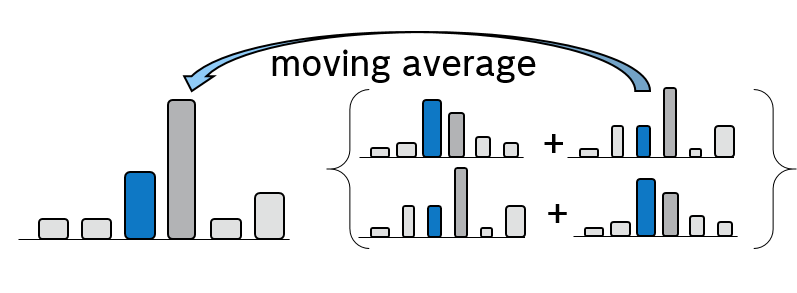}
	\caption{We filter training samples based on the moving average (mva-) predictions (left) of predictions from previous training iterations (right). Our strategy is extremely efficient since no additional model inference step is required. It is also effective since predictions of noisy samples tend to fluctuate during training.}
	\label{fig:mva-filtering}
\end{figure}
The model's predictions required for filtering can be stored during training directly. However, the predictions for noisy samples tend to fluctuate. For example, take a cat wrongly labeled as a tiger.
Other cat samples would encourage the model to predict the given cat image as a cat.  Contrary, the wrong label \emph{tiger} regularly pulls the model back to predict the \emph{cat} as a tiger. Hence, using the model's predictions gathered in one single training epoch for filtering is sub-optimal.

Instead, we propose to collect the sample predictions over multiple training epochs. This scheme is displayed in Fig.~\ref{fig:mva-filtering}. For each sample, we store the moving averaged predictions, accumulated over the last iterations. Besides having a more stable basis for the filtering step, our proposed procedure also leads to negligible memory and computation overhead.

Due to continuous training of the best model from the previous model, computation time can be significantly reduced, compared to re-training the model from scratch. On the new filtered dataset, the model must only slowly adapt to the new noise ratio contained in the training set. Depending on the computation budget, a maximal number of iterations for filtering can be set to save time.  

Moreover, the new training procedure does not require specific mechanisms or algorithms which need to be implemented or fine-tuned. Implementation-wise, it can be realized by looping the standard training procedure and filter potentially noisy samples at the end of each training run.

\subsection{Unsupervised learning to counter label noise}
\label{s:unsupervised}

Although the proposed learning procedure is not restricted to classification tasks, in this work, we explain the procedure for classification as a use-case.

Model training is performed using two types of learning objectives: (1) supervised and (2) unsupervised losses. Supervised learning from noisy-labeled samples is straightforward and can be done with typical n-way-classification losses. The unsupervised learning objective, however, requires a design choice of which data to be used (defined in Section~\ref{ss:general scheme})  and how to learn from them.
\subsubsection{Learning from unlabeled data }
We learn from \emph{all} data points in a semi-supervised fashion. Concretely, in addition to supervised learning with filtered labels, unsupervised learning is applied to the entire dataset. Our learning strategy can take advantage of unsupervised learning from a large dataset, and therefore it has a potentially large regularization effect against label noise. Unsupervised learning objectives impose additional constraints on all samples, which are hard to follow for wrongly labeled samples. These constraints could be a preference of extreme predictions (Entropy-loss) or  non-fluctuating model predictions over many past iterations (Mean-teacher-loss). Both constraints are explained in the following.

\paragraph{Entropy minimization}
The typical entropy loss for semi-supervised learning is shown in Fig.~\ref{fig:lossess}. It encourages the model to provide extreme predictions (such as $0$ or $1$) for each sample. Over a large number of samples, the model should balance its predictions over all classes. 
\begin{figure}[t]
	\centering
	
	\begin{subfigure}{.21\textwidth}
		\begin{center}
			\includegraphics[width=1\linewidth]{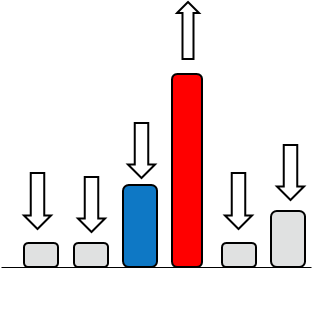}
			\caption{}
			\label{fig:noise-recall}
		\end{center}
	\end{subfigure}            
	\begin{subfigure}{.21\textwidth}
		\begin{center}
			\includegraphics[width=.8\linewidth]{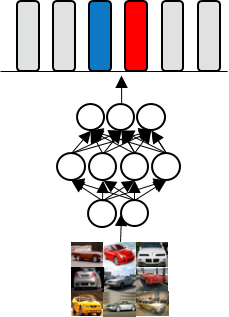}
			\caption{}
			\label{fig:noise-recall}
		\end{center}
	\end{subfigure}            
	
	\caption{The entropy loss for semi-supervised learning. (a) Extreme predictions such as $[0,1]$ are encouraged by minimizing the entropy on each prediction. (b) Additionally, maximizing the entropy of the mean prediction on the entire dataset or a large batch forces the model to balance its predictions over multiple samples.}
	\label{fig:lossess}
\end{figure}

The entropy loss can easily be applied to all samples to express the uncertainty about the provided labels. Alternatively, the loss can be combined with a strict filtering strategy, as in our work, which removes the labels of potentially wrongly labeled samples.

For a large noise ratio, predictions of wrongly labeled samples fluctuate strongly over previous training iterations. Amplifying these network decisions could lead to even noisier models model. Combined with  iterative filtering, the framework will have to rely on a single noisy model snapshot. In the case of an unsuitable snapshot, the filtering step will make many wrong decisions. 

\paragraph{Mean Teacher model}
A better way to perform semi-supervised learning and counteract label noise is to employ the Mean Teacher model~\cite{tarvainen2017mean}. The Mean Teacher model follows the student-teacher learning procedure from~\cite{hinton2015distilling}. The main idea is to create a virtuous learning cycle, in which the student continually learns to surpass the (better) teacher. Concretely, the Mean Teacher is an exponential moving average of the student models over training iterations. 

\begin{figure}
	\centering
	\includegraphics[width=.8\linewidth]{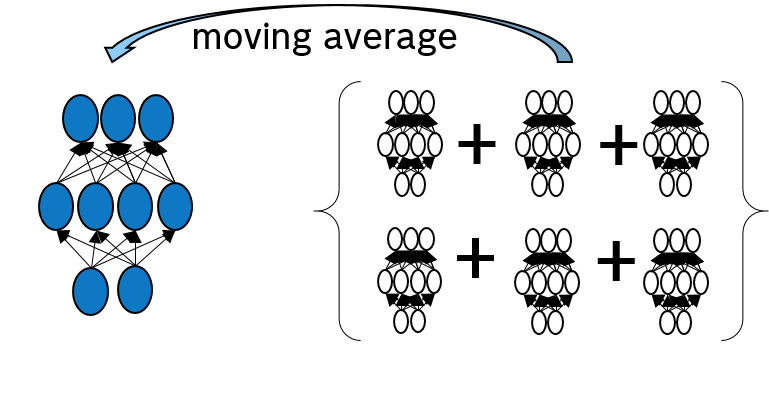}
	\caption{Mean Teacher (left)  is a moving average of the weights of student models from previous training iterations (right). In the presence of large label noise, single student models fluctuate strongly, due to the wrong learning signal. Having a temporal ensemble of model weights counteracts this effect and provides significantly better and more stable predictions. }
	\label{fig: Mean Teacher weight averaging }
\end{figure}

In contrast to learning from the entropy-loss, the Mean-Teacher solves precisely the problem of noisy models snapshots. The teacher-model is a moving-average from the past training iterations and hence much more stable than a single snapshot. The training of such a model is shown in Fig.~\ref{fig: Mean Teacher weight averaging }

\paragraph{Mean Teacher model for iterative filtering}
Given the setting in Section~\ref{ss:general scheme}, we apply the Mean Teacher algorithm in each iteration $i$ in the $train\_and\_valid({\mathcal D}_{i}, {\mathcal D}_{val})$ procedure as follows.
\begin{itemize}
	\setlength\itemsep{0em}
	\item Input: examples with potentially clean labels $D_i^{clean}$ from the filtering procedure. In the beginning ($i=0$), $D_0^{clean}= D_0$
	\item Initialize a supervised neural network as the student model $M^s_i$.
	\item Initialize the Mean Teacher model $M^t_i$ as a copy of the student model with all weights detached. 
	\item Let the loss function be the sum of normal classification loss of $M^s_i$ and the \emph{consistency loss} between the outputs of $M^t_i$ and $M^t_i$
	\item Select an optimizer
	\item In each training iteration: 
	
	\begin{itemize}
		\setlength\itemsep{0em}
		\item Update the weights of $M^s_i$ using the selected optimizer 
		\item Update the weights of $M^t_i$ as an exponential moving average of the student weights  
		\item Evaluate performance of $M^s_i$ and $M^t_i$ over ${\mathcal D}_{val}$ to verify the early stopping criteria.   
	\end{itemize}
	\item Return the best $M^t_i$
\end{itemize}

The \textit{consistency loss} between students and teachers output distribution can be realized with Mean-Square-Error or Kullback-Leibler-divergence.
\paragraph{Overlapping data split between labeled and unlabeled samples}
While traditionally the dataset is strictly divided into non-overlapping labeled and unlabeled sets, we treat all samples also as unsupervised samples, even if they are in the set of filtered, labeled samples. 

This is important since despite the filtering the provided labels can be wrong. By considering them additionally as unsupervised samples, the consistency of the model prediction for a potentially noisy sample is evaluated among many other samples, resulting in more consistent model predictions. Therefore, learning from all samples in an unsupervised fashion provides a stronger regularization effect against label noise. 

\section{Related Works}

Different approaches to counter label noise have been proposed in~\cite{azadi2015auxiliary, reed2014training, ren_learning_2018-3, jiang_mentornet:_2017-3, Jenni2018DeepBL}. Some of these works~\cite{azadi2015auxiliary, ren_learning_2018-3} require additional clean training data. Often, the loss for potentially noisy labels is re-weighted softly to push the model away from the wrong label~\cite{jiang_mentornet:_2017-3, ren_learning_2018-3}. 

Compared to these works, we perform an \emph{extreme} filtering by setting the sample weight of the potentially wrongly labeled samples to $0$. These labels are no longer used for the supervised objective of the task. Moreover, we perform the filtering step very seldom, in contrast to epoch-wise-samples re-weighting of previous approaches. Furthermore, contrary to all previous robust learning approaches, we utilize iterative training \emph{combined with semi-supervised learning} to combat label noise for the first time.

Despite recent advances in semi-supervised learning~\cite{rasmus2015semi,makhzani2015adversarial,kingma2014semi,kumar2017semi, springenberg2015unsupervised, miyato2018virtual,dai2017good}, it has not been considered as a regularization technique against label noise. Semi-supervised learning often uses generative modeling~\cite{kingma2013auto,kingma2016improved, rezende2014stochastic, goodfellow2014generative} as an auxiliary task.
In contrast to using generative models, the \emph{Mean Teacher} model proposed in~\cite{tarvainen2017mean} has a more stable training procedure. The Mean Teacher does not require any additional generative model. 
More details are explained  in Section~\ref{s:unsupervised}.

Typically, unsupervised learning is only applied to unlabeled data. Contrary, in our approach, unsupervised learning is applied to \emph{all samples} to expresses the uncertainty of the provided labels.

Although previous robust learning approaches such as ~\cite{wang2018iterative} also use iterative training and filtering, their approach does not employ learning from removed samples in an unsupervised fashion. Furthermore, they always filter strictly, i.e., each sample removal decision is final. 

In IF-SSL we only filter potentially noisy labels from the original label set, but still, use the corresponding instances for unsupervised learning. This gives the model a chance to revert a wrong filtering decision in earlier iterations.  

Further, our framework is intentionally kept more simple and generic than previous techniques. The focus of our framework is the iterative filtering of noisy labels while learning from all samples in an unsupervised fashion as a form of regularization. This paradigm is hence easily transferable to other tasks than classification.

\section{Evaluation}

\subsection{Description of Experiments}
\subsubsection{Tasks} 
\begin{table}[t]
	\caption{Dataset description. Classification tasks on CIFAR-10 and CIFAR-100 with uniform noise. Note that the noise on the training and validation set is not correlated. Hence, maximizing the accuracy on the noisy set provides a useful (but noisy) estimate for the  generalization ability on unseen test data.}
	\vskip 0.15in
	\begin{center}
		\begin{small}
			\begin{sc} 
				\centering
				\begin{tabular}{ll*{2}{p{1.3cm}}}
					\toprule
					\midrule
					&Type &CIFAR-10  & CIFAR-100\\
					\cmidrule{2-4}
					Task & classification & 10-way & 100-way \\
					Resolution &\multicolumn{3}{c}{32x32} \\
					\cmidrule{2-4}
					\multirow{3}{*}{Data} &  Train (noisy) &  45000 &45000 \\
					&  Valid (noisy)&  5000  &5000 \\
					&  Test (clean)&    10000 &10000 \\
					\midrule
					\bottomrule
				\end{tabular} 
				\label{Tab: Dataset}
			\end{sc}
		\end{small}
	\end{center}
\end{table}
We evaluate our approaches to noisy CIFAR-10 and CIFAR-100 with different label noise ratios as shown in Tab.~\ref{Tab: Dataset} . We analyze the typical situation with uniform noise, in which a label is randomly flipped to another class. Further experiments on ImageNet-ILSVRC is in the Appendix.

\subsubsection{Comparisons to related works}
We compare our framework IF-SSL (Iterative Filtering + Semi-supervised Learning) to previous robust learning approaches such as MentorNet~\cite{jiang_mentornet:_2017-3}, Learned and random sample weights from~\cite{ren_learning_2018-3}, S-Model~\cite{goldberger2016training},  bi-level learning~\cite{Jenni2018DeepBL}, Reed-Hard~\cite{reed2014training} and Iterative learning in open-set problems~\cite{wang2018iterative}.  

Hyperparameters and early-stopping are determined on the noisy validation set. This is possible because the noise of the validation and training sets is not correlated. Hence, higher validation performance often results in superior test performance. 

Additionally,~\cite{ren_learning_2018-3} considered the setting of having a small clean validation set of 1000 images. For comparison purposes, we also experiment with a small clean set for early stopping.

Whenever possible, we adopt the performances of their methods from the corresponding publications. Sometimes, not all numbers are reported in these publications.

\subsubsection{Network configuration and training}
For the basic training of semi-supervised models, we use a Mean Teacher model~\cite{tarvainen2017mean} available on GitHub \footnote{https://github.com/CuriousAI/mean-teacher}. The students and teacher networks are residual networks~\cite{he2016deep} with 26 layers. They are trained with Shake-Shake-regularization~\cite{gastaldi2017shake}. We use the PyTorch~\cite{paszke2017automatic} implementation of the network and keep the training settings close to~\cite{tarvainen2017mean}. The network is trained with Stochastic Gradient Descent. In each filtering iteration, the model is trained for a maximum of $300$ epochs, with a patience of $50$ epochs. For more training details, see the appendix.

To filter the noise iteratively, we use the early stopping strategy based on the  validation set. After the best model is found, we use it to filter out potentially noisy samples from the noisy training label set at iteration $0$. In the next iteration, the previously best model is fine-tuned on the new dataset. All data is used for unsupervised learning, while supervised learning only considers the filtered labels set at the current iteration. We stop the iterative filtering if no better model is found. 

\subsubsection{ Structure of analysis}
We start with the analysis of our model's performance under different noise ratios. We compare our performance to other previously reported approaches in learning under different noise ratios using the accuracy metric on CIFAR-10 and CIFAR-100. The subsequent ablation study highlights the importance of each component in our framework.

Further, we analyze the consequence of applying our iterative filtering scheme to different network architectures. Afterwards, we show the performance of simple unsupervised learning objectives, with and without our iterative filtering scheme. For more experiments, we refer to the supplemental material. 

\subsection{Robust Learning Performance Evaluation}
\subsubsection{Model accuracy under label noise}

\begin{figure*}[t]
	\begin{subfigure}{.33\textwidth}
		\begin{center}
			\includegraphics[width=1\linewidth]{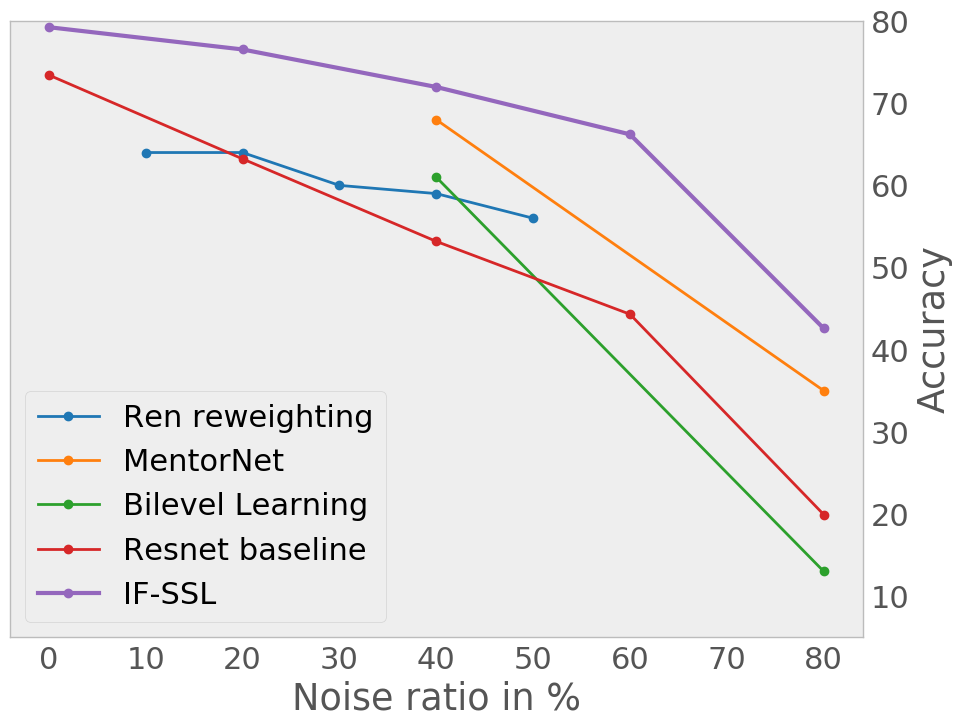}
			\caption{Accurracy on CIFAR-100}
			\label{fig:cifar-100:acc}
		\end{center}
	\end{subfigure}
	\begin{subfigure}{.33\textwidth}
		\begin{center}
			\includegraphics[width=1\linewidth]{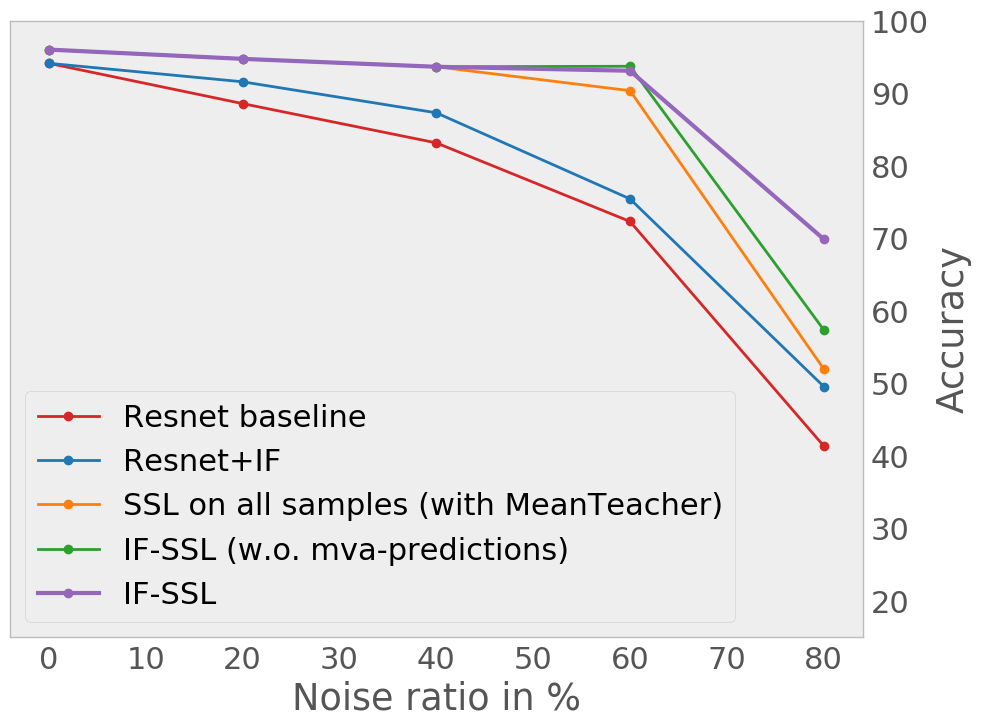}
			\caption{Ablation exps. on CIFAR-10}
			\label{fig:noise-perc}
		\end{center}
	\end{subfigure}
	\begin{subfigure}{.33\textwidth}
		\begin{center}
			\includegraphics[width=1\linewidth]{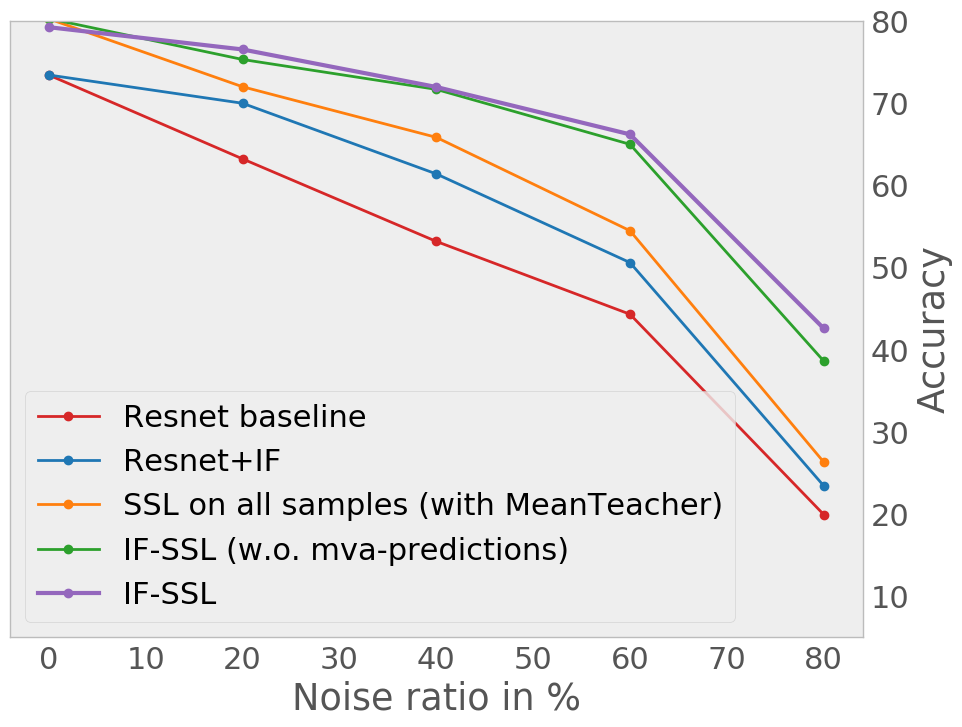}
			\caption{Ablation exps. on CIFAR-100}          
			\label{fig:noise-recall}
		\end{center}
	\end{subfigure}

	\caption{(a) Noisy CIFAR-100: Model performance measured in accuracy for learning under different label noise ratio. IF-SSL retains high accuracy despite the high noise ratio. Contrary, the approach from~\cite{ren_learning_2018-3} under-performs even in case of small label noise. See Fig.~\ref{fig:cifar-10:acc} for Cifar-10. (b-c)Ablation study: the importance of iterative filtering and semi-supervised learning for robust learning. Performance is measured on Cifar-10 (b) and Cifar-100 (c)
		Our IF-SSL provides a solid baseline. By using the moving average predictions of samples for filtering, the baseline can even be improved further. This technique is especially useful in the presence of large label noise.}
	\label{fig:ablation}
\end{figure*}    

\begin{table}[t]
	\centering
	\caption{Classification accuracy in \% of robust learning models on CIFAR-10 annd CIFAR-100 with 40\% and 80\% label noise.  Random weights and learned weights indicate weights assigning strategy in~\cite{ren_learning_2018-3}. (\mbox{*}) denotes approaches, which utilizes  extra set of 1000 clean samples~\cite{ren_learning_2018-3}. For a fair comparison, we compare our approach IF-SSL using (1) noisy validation set and (2) 1000 clean samples for early stopping.
		In all cases, our IF-SSL outperforms previous techniques regarding accuracy and robust learning behavior by a large margin. Having a small data-set for validation increases our performances slightly.}
	\vskip 0.15in 
	\begin{small}
		\begin{sc}
			\begin{tabular}{*{1}{p{3.15cm}} *{4}{p{.75cm}}}
				\toprule
				\midrule
				
				& \multicolumn{2}{c}{Cifar-10} & \multicolumn{2}{c}{CIFAR-100} \\
				Noise ratio    & 40\% & 80 \%  & 40\% & 80 \% \\
				
				\midrule
				\multicolumn{5}{c}{USING NOISY DATASET ONLY} \\
				\midrule
				Reed-Hard~\cite{reed2014training} & 69.66 & - & 51.34 & - \\
				S-model~\cite{goldberger2016training} & 70.64   & -  & 49.10 & - \\
				\cite{wang2018iterative} & 78.15& - & - & - \\
				Rand. weights~\cite{ren_learning_2018-3} &86.06  & - &58.01  & -\\

				Bi-level-model~\cite{Jenni2018DeepBL} & 89& 20 & 61 & 13  \\
				MentorNet~\cite{jiang_mentornet:_2017-3} & 89  & 49 & 68 & 35\\
				
				Resnet26 baseline& 83.2   & 41.37 & 53.18 & 19.12\\
				(Ours) IF+SSL & \textbf{93.7}  & \textbf{69.91} & \textbf{71.98} & \textbf{42.09}\\
				\midrule
				\multicolumn{5}{c}{USING 1000 CLEAN IMAGES} \\
				\midrule
				Mentornet~\cite{jiang_mentornet:_2017-3}* & 78 &  -  &59  & - \\
				Rand.  weights~\cite{ren_learning_2018-3}* &  86.55    & - &58.34  & -\\
				Ren et al~\cite{ren_learning_2018-3}* &  86.92    & - & 61.31  & - \\
				(Ours) IF+SSL* & \textbf{95.1}  &  \textbf{79.93}  & \textbf{74.76} &  \textbf{46.43} \\
				\midrule
				\bottomrule
			\end{tabular}
			\label{tab:all accuracies}
		\end{sc}
	\end{small}
\end{table}

Results for typical scenarios with noise ratio of 40\% or 80\% on CIFAR-10 and CIFAR-100 are shown in Tab.~\ref{tab:all accuracies}. More results are visualized in Fig.~\ref{fig:cifar-10:acc} (CIFAR-10) and Fig.~\ref{fig:cifar-100:acc} (CIFAR-100). The baseline model is the typical ResNet-26 with a n-way-classification loss (Negative-Log-likelihood-objective).

Compared to the model baseline and other previously reported approaches, IF-SSL outperforms them by a large margin. Even in areas of high noise ratio up to 80\%, the classification performance of our model remains highly robust. 
Despite the noisy validation set, our model still identifies the noisy labels and filters them out. On CIFAR-10 and CIFAR-100, our model IF-SSL achieves 20\% and 7\% absolute improvement over previously reported results. 

A small clean validation set gives the model an even better estimate of the generalization error on unseen data (IF-SSL*). Due to the iterative filtering scheme, our model always attempts to improve the performance on the validation set as much as possible, without doing gradient steps on it. At the time of convergence, the model \emph{always} has a loss very close to $0$. Contrary, to prevent over-fitting, a simple early stopping scheme usually leads to a high remaining training loss. Our filtering framework indicates that it is meaningful to learn further from easy samples and to treat the other samples as unlabeled. See the appendix for training visualizations. 

Previous works utilize strict filtering, where removed samples are not re-considered in later filtering iterations, whereas iterative filtering always filters based on the provided label set at iteration $0$. The experiments show the enormous benefit of this. The IF-SSL* using clean validation set only achieves 70.93 \% at 80\% noise when the samples are completely removed. The improvement also stagnates after one single filtering iteration. Hence, for a fair comparison with all filtering baselines, we always use the filtered data as unlabeled  samples if not stated otherwise.
More details and experiments can be found in the appendix.

\subsubsection{Ablation Study} 
\begin{table}[t]
	\centering
	\caption{Ablation studies. (-X) means IF-SSL without feature X.  
		Note that training with iterative filtering or semi-supervised learning only results in similar performance. Combining both techniques leads to a very strong baseline. However, our proposed filtering based on moving-average predictions even increase the performance further in case of large label noise (up to 80\%).}
	\vskip 0.15in 
	\begin{small}
		\begin{sc}
			\begin{tabular}{*{1}{p{3cm}} *{4}{p{.8cm}}}
				\toprule
				\midrule
				& \multicolumn{2}{c}{Cifar-10} & \multicolumn{2}{c}{CIFAR-100} \\
				noise ratio & 40\% & 80 \%  & 40\% & 80 \% \\
				\midrule
				(Ours) IF+SSL & \textbf{93.7}  & \textbf{69.91} & \textbf{71.98} & \textbf{42.09}\\
				\midrule
				- mva-predictions  & 93.77 & 57.4 & 71.69 & 38,61\\
				SSL-only & 93.7 & 52.5 & 65.85 & 26.31 \\ 
				IF & 87.35 & 49.58&  61.4  & 23.42\\
				Resnet26 & 83.2  & 41.37 & 53.18 & 19.92 \\ 
				\midrule
				\bottomrule
			\end{tabular}
			\label{tab:Ablation studies}
		\end{sc}
	\end{small}
\end{table}

Tab.~\ref{tab:Ablation studies} indicates the importance of the iterative filtering and semi-supervised learning procedure in our framework. Performing semi-supervised learning (on all samples) or iterative filtering alone leads to similar performances. When combined (IF-SSL
without moving-average-predictions), the model is highly robust at 40\% noise. 

With a higher noise ratio of 80\% however, the model's predictions on training samples fluctuate strongly. Hence, merely taking the model's predictions at one specific epoch leads to a sub-optimal filtering step. Contrary, our approach IF-SSL proposes to utilize moving-average predictions which are significantly more stable. Compared to the baseline IF-SSL without moving-average predictions, this technique leads to 12\% and 3.5 \% absolute improvement on CIFAR-10 and CIFAR-100 respectively.

Naive training or leaving out any of the proposed mechanism leads to rapid performance decrease. Our framework combines the strength of both techniques to form an extremely effective regularizer against learning from label noise.

\subsubsection{Iterative filtering with different architectures}

Tab.~\ref{tab: different architectures} shows the effect of iterative filtering on various architectures. For traditional network training, Resnet26 performs best and slightly better than its shallower counterpart Resnet18. Extremely deep architectures like Resnet101 suffer more from the high-noise ratios.

\begin{table}[t]
	\centering
	\caption{Performance of different architectures on Cifar-10 and CIFAR-100. Best performance in the respective category is marked. Iterative filtering always increases the accuracy on the test set by a large margin. Our IF-SSL combines the strength of iterative filtering with semi-supervised learning to counter label noise much more effectively. }
	\vskip 0.15in 
	\begin{small}
		\begin{sc}
			\begin{tabular}{l cccc}
				\toprule
				\midrule
				
				& \multicolumn{2}{c}{Cifar-10} & \multicolumn{2}{c}{CIFAR-100} \\
				Noise ratio    & 40\% & 80 \%  & 40\% & 80 \% \\
				
				\midrule
				
				ResNet18 & 75.03  & 34.9 & 43.34 & 5.34  \\
				ResNet26 & \textbf{83.2}  & \textbf{41.37} & \textbf{53.18} & \textbf{19.92}  \\
				ResNet101 & 68.14  & 32.5 & 36.02 & 13.24  \\
				
				\midrule
				\multicolumn{5}{c}{With Iterative Filtering} \\
				\midrule
				
				ResNet18-IF & 85.75  & 42.84&  57.86 & 21.27\\
				ResNet26-IF & \textbf{87.35} & \textbf{49.58}&  \textbf{61.4}  & \textbf{23.42}\\
				ResNet101-IF & 82.46  & 33.2&  47.11 & 6.50\\
				\midrule
				
				(Ours) IF+SSL & \textbf{93.7}  & \textbf{69.91} & \textbf{71.98} & \textbf{42.09}  \\
				\midrule
				\bottomrule
			\end{tabular}
			\label{tab: different architectures}
		\end{sc}
	\end{small}
\end{table}

With the proposed iterative filtering, the performance gaps between different models are massively reduced. After iterative filtering, Resnet26 and Resnet18 perform similarly well and provide a very strong baseline. IF-SSL achieves up to 19\% absolute improvement over the best Resnet26+IF-baseline at 80\% noise ratio.

\subsubsection{Semi-supervised learning techniques + iterative filtering}

\begin{table}[t]
	\centering
	\caption{Analysis of semi-supervised learning strategies. Push-away-loss, entropy-loss, and mean-teacher-loss are separately considered. The best two performances from the respective category are marked. The entropy loss is further applied to \emph{all} training samples or only to the removed samples from earlier filtering steps.}
	\vskip 0.15in 
	\begin{small}
		\begin{sc}
			\begin{tabular}{l*{2}{p{.7cm}}}
				
				\toprule
				\midrule
				
				& \multicolumn{2}{c}{Cifar-10}  \\
				Noise ratio & 40\% & 80 \% \\
				\midrule
				
				Resnet26  & 83.2 & 41.37 \\ 
				Entropy (all-samples) & \textbf{85.98} & \textbf{46.93} \\ 
				Mean Teacher (all-samples) & \textbf{90.4} & \textbf{52.5}  \\ 
				
				\midrule
				\multicolumn{3}{c}{With Iterative Filtering} \\
				\midrule
				Push Away+IF  & \textbf{90.47} & 50.79 \\
				Entropy (all samples)+IF  & 90.4 & 52.46 \\     
				Entropy (unlabeled samples)+IF& 90.02 & \textbf{53.44} \\ 
				Mean Teacher + IF (ours) & \textbf{93.7}  & \textbf{69.91} \\
				
				\midrule
				\bottomrule
			\end{tabular}
			\label{tab: different SSL strategy+IF}
		\end{sc}
	\end{small}
\end{table}

Tab.~\ref{tab: different SSL strategy+IF} shows different semi-supervised learning strategies with and without iterative filtering.
The push-away-loss corresponds to assigning negative weights to potentially noisy labels. The entropy loss minimizes the network's uncertainty on a set of samples. Since our labels are all potentially noisy, it is meaningful to apply this loss to \emph{all} training samples instead of removed samples only. Hence we compare both variants. The Mean-teacher loss is always applied to \emph{all} samples (details in the appendix).

Without filtering: Learning from the entropy-loss performs second-best, when the uncertainty is minimized on all samples. Without the previous filtering step, there is no set of unlabeled samples to perform a traditional semi-supervised-learning.
The Mean-teacher performs best since the teacher represents a  stable model state, aggregated over multiple iterations.

With filtering: Applying entropy-loss to all samples or only unsupervised samples leads to very similar performance. Both are better than the standard push-away-loss. Our Mean Teacher achieves by far the best performance, due to the temporal ensemble of models and sample predictions for filtering.

\section{Conclusion}

In this work, we propose a training pipeline for robust learning. Our method relies on two key components: (1) iterative filtering of potentially noisy labels, and (2) regularization by learning from all raw data samples in an unsupervised fashion.

We have shown that neither iterative noise filtering (IF) nor semi-supervised learning (SSL) alone is sufficient to achieve competitive performance. Contrary, we combine IF and SSL and extend them with crucial novel components for more robust learning.

Unlike previous filtering approaches, we always filter the initial label set provided at the beginning. Furthermore, we utilize a temporal ensemble of model predictions as the basis for the filtering step.

The proposed algorithm is evaluated on classification tasks for CIFAR-10 and CIFAR-100 with a varying label noise ratio from 0\% to 80\%. We show results both for a clean validation set and a noisy one. In both cases, we show that using the filtered data as unlabeled samples significantly outperforms complete removal of the data. As a consequence, the proposed model consistently outperforms state of the art at all levels of label noise. Despite the simplicity of the training pipeline, our approach shows robust performance even in case of high noise ratios.

\bibliography{iclr2018_conference}

\bibliographystyle{icml2019}

\appendix
\section{Large-scale classification on ImageNet-ILSVRC-2015}

\begin{table}[t]
	\centering
	\caption{ImageNet experiments (40\%-noise). All performances are measured in \% accuracy. Performances are reported on the clean validation set. The basic model indicates general network training. Mentornet was the best previously reported results. SSL (semi-supervised-learning) means that all samples are treated additionally as unlabeled for the unsupervised task. IF-SSL is our framework, which uses iterative filtering and SSL. Our model is more robust compared to previous works, despite the difficulty of the ImageNet dataset. The best robust learning method for each architecture is marked.
		Note that   Mentornet\mbox{*}  is based on Resnet-101. We chose the weaker model Resnext50 (and ResNext18) to reduce the run-time.}
	\vskip 0.15in 
	\begin{small}
		\begin{sc}
			\begin{tabular}{l cccc}
				\toprule
				\midrule    & \multicolumn{2}{c}{Resnext18} &\multicolumn{2}{c}{Resnext50} \\
				Accurracy   & P@1 & P@5 & P@1 & P@5  \\
				\midrule
				Mentornet* &  - & - & 65.10 & 85.90 \\ 
				Basic model &50.6 & 75.99&56.25 & 80.90\\
				SSL         &58.04 & 81.82& 62.96& 85.72\\ 
				IF-SSL (Ours)  &\textbf{66.92} & \textbf{86.65}& \textbf{71.31}  & \textbf{89.92} \\
				\midrule
				\bottomrule
			\end{tabular}
			\label{tab: Filtering stragies}
		\end{sc}
	\end{small}
\end{table}

Tab. ~\ref{tab: Filtering stragies} shows the precision@1 and @5 of various models, given 40\% label noise in the training set. Our networks are based on ResNext18 and Resnext50. Note that MentorNet \cite{jiang_mentornet:_2017-3} uses Resnet101 (P@1:78.25) \cite{goyal2017accurate}, which has similar performance compared to Resnext50 (P@1: 77.8)\cite{xie2017aggregated} on the standard ImageNet validation set. Although Resnext50 is a weaker model, we opt for the Resnext counterparts because of the significantly shorter training time. Hence, our performance reported with ResNext50 is a lower-bound of our approach with Resnet-101. Results with Resnext18 and Resnext50 indicates, that stronger models results in higher accuracy in our framework.

Despite the weaker model, IF-SSL (ResNext50) surpasses the best previously reported results by more than 5\% absolute improvement. Even the significantly weaker model ResNext18 outperforms MentorNet based on a very powerful ResNet101 network.

\section{Complete removal of samples}
\begin{table}[t]
	\centering
	\caption{Accuracy of the complete removal of samples during iterative filtering on CIFAR-10 and CIFAR-100. The underlying model is the MeanTeacher based on Resnet26. When samples are completely removed from the training set, they are no longer used for either supervised-or-unsupervised learning.  This common strategy from previous works leads to rapid performance breakdown.
	}
	\vskip 0.15in 
	\begin{small}
		\begin{sc}
			\begin{tabular}{l cccc}
				\toprule
				\midrule    & \multicolumn{2}{c}{Cifar-10} & \multicolumn{2}{c}{CIFAR-100} \\
				Noise ratio    & 40\% & 80 \%  & 40\% & 80 \% \\
				\midrule
				\multicolumn{5}{c}{Using noisy data only} \\
				\midrule
				
				Compl. Removal& 93.4  & 59.98 & 68.99 & 35.53  \\
				IF-SSL (Ours)  & \textbf{93.7}  & \textbf{69.91} & \textbf{71.98} & \textbf{42.09}  \\
				\midrule
				\multicolumn{5}{c}{With clean validation set} \\
				\midrule
				\midrule
				Compl. Removal& 94.39  & 70.93 & 71.86 & 36.61  \\
				IF-SSL (ours) & \textbf{95.1}  &  \textbf{79.93}  & \textbf{74.76} &  \textbf{46.43} \\
				\midrule
				\bottomrule
			\end{tabular}
			\label{tab: Filtering stragies}
		\end{sc}
	\end{small}
\end{table}

Tab. \ref{tab: Filtering stragies} shows the results of deleting samples from the training set. It leads to large performances gaps compared to our strategy (IF-SSL), which considers the removed samples as unlabeled data. In case of a considerable label noise of 80\%, the gap is close to 9\%.

Continuously using the filtered samples lead to significantly better results. The unsupervised-loss provides meaningful learning signals, which should be used for better model training.

\section{Training process}

\begin{figure}[t]
	\centering
	
	\begin{subfigure}{.45\textwidth}
		\begin{center}
			\includegraphics[width=1\linewidth]{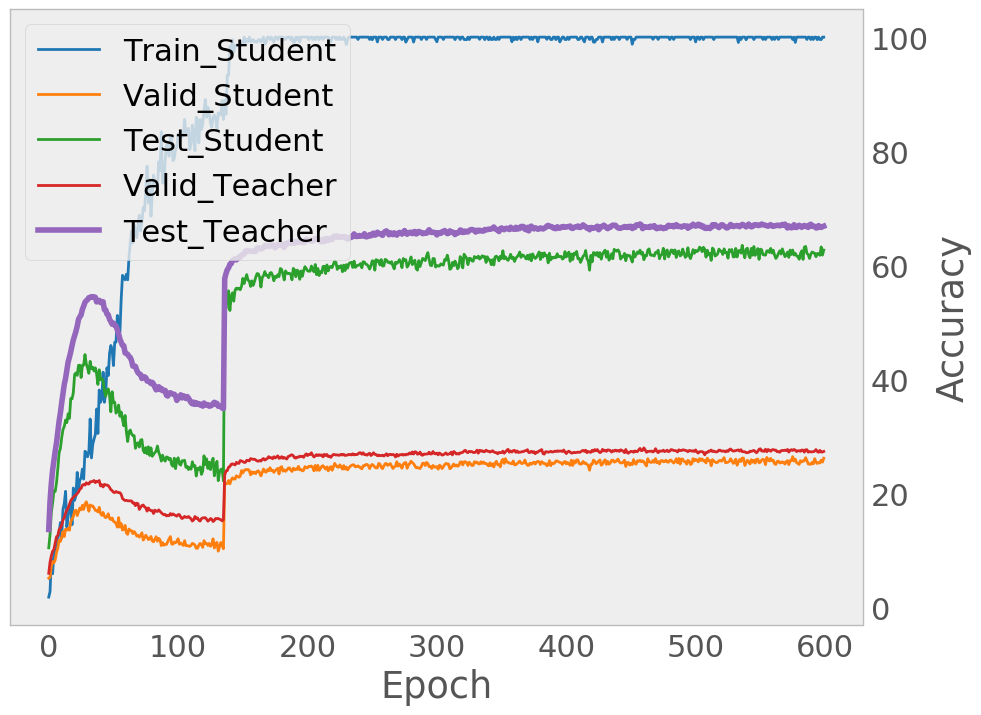}
			\caption{}
			\label{fig:noise-perc}
		\end{center}
	\end{subfigure}
	\begin{subfigure}{.45\textwidth}
		\begin{center}
			\includegraphics[width=1\linewidth]{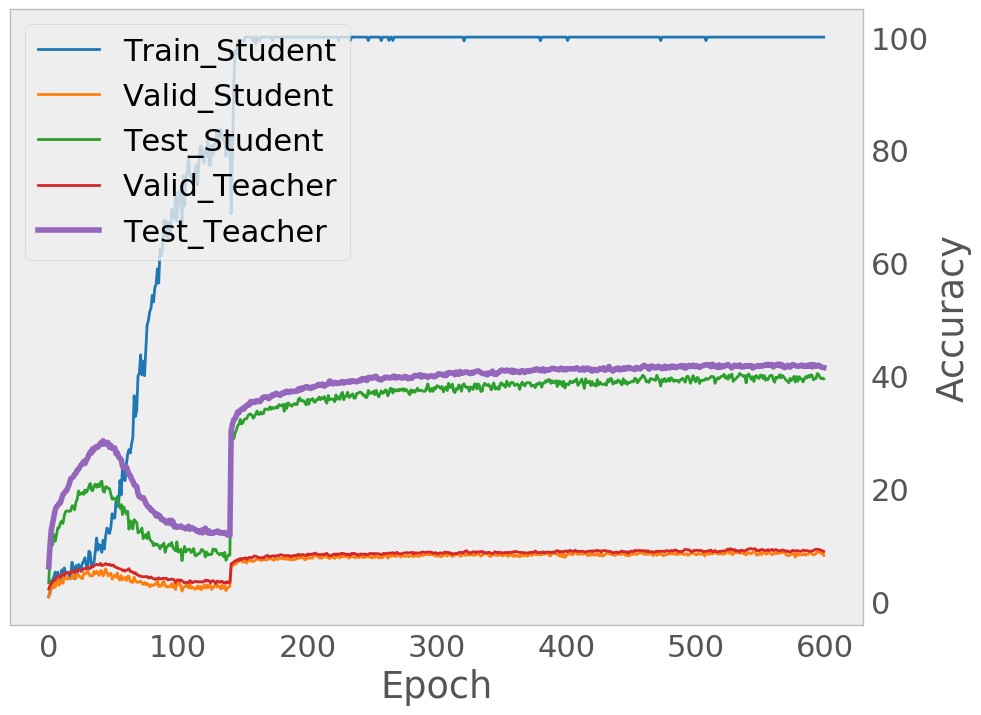}
			\caption{}
			\label{fig:noise-perc}
		\end{center}
	\end{subfigure}
	\caption{Sample training curves of our approach IF-SSL on CIFAR-100 with (a) 60\% and (b) 80\% noise, using noisy validation data. Note that with our approach, the training loss remains close to $0$. Further, note that the mean-teacher continously outperforms the noisy student models. This shows the positive effect of temporal emsembling to counter label noise.}
	\label{fig:training curves}
\end{figure}

Fig. ~\ref{fig:training curves} shows the sample training processes of IF-SSL under 60\% and 80\% noise on CIFAR-100. The mean-teacher always outperform the student models. Further, note that regular training leads to rapid over-fitting to label noise. 

Contrary, with our effective filtering strategy, both models slowly increase their performance while the training accuracy approaches 100\%. Hence, by using iterative filtering, our model could erase the inconsistency in the provided labels set.

\section{Training details}
\subsection{CIFAR-10 and CIFAR-100}
\paragraph{Network training}
For the training our model IF-SSL, we use the standard configuration provided by \cite{tarvainen2017mean} \footnote{https://github.com/CuriousAI/mean-teacher}. Concretely, we use the SGD-optimizer with Nesterov \cite{sutskever2013importance} momentum, a learning rate of 0.05 with cosine learning rate annealing \cite{loshchilov2016sgdr}, a weight decay of 2e-4,  max iteration per filtering step of 300, patience of 50 epochs, total epochs count of 600. 

For basic training of baselines models without semi-supervised learning, we had to set the learning rate to 0.01. In the case of higher learning rates, the loss typically explodes. Every other option is kept the same.
\paragraph{Semi-supervised learning}
For the mean teacher training, additional hyperparameters are required. In both cases of CIFAR-10 and CIFAR-100, we again take the standard configuration with the consistency loss to mean-squared-error and a consistency weight: 100.0, logit distance cost: 0.01, consistency-ramp-up:5. The total batch-size is 512, with 124 samples being reserved for labeled samples, 388 for unlabeled data. Each epoch is defined as a complete processing of all unlabeled data. 
When training without semi-supervised-learning, the entire batch is used for labeled data.
\paragraph{Data augmentation}
The data are normalized to zero-mean and standard-variance of one. Further, we use real-time data augmentation with random translation and reflection, subsequently random horizontal flip. The standard PyTorch-library provides these transformations.
\subsection{ImageNet-ILSVRC-2015}
\paragraph{Network Training}

The network used for evaluation were ResNet \cite{he2016deep} and Resnext \cite{xie2017aggregated} for training. All ResNext variants use a cardinality of 32 and base width of 4 (32x4d). ResNext models follow the same structure as their Resnet counterparts, except for the cardinality and base width. 

All other configurations are kept as close as possible to \cite{tarvainen2017mean}. The initial learning rate to handle large batches \cite{goyal2017accurate} is set to 0.1; the base learning rate is 0.025 with a single cycle of cosine annealing. 

\paragraph{Semi-supervised learning}
Due to the large images, the batch size is set to 40 in total with 20/20 for labeled and unlabeled samples respectively. We found the Kullback-divergence leads to no meaningful network training. Hence, we set the consistency loss to mean-squared-error, with a weight of 1000. We use consistency ramp up of 5 epochs to give the mean teacher more time in the beginning. Weight decay is set to 5e-5; patience is four epochs to stop training in the current filtering iteration.
\paragraph{Filtering}
We filter noisy samples with the topk=5 strategy, instead of topk=1 as on CIFAR-10 and CIFAR-100. That means the samples are kept for supervised training if their provided label lies within the top 5 predictions of the model.
The main reason is that each image of ImageNet might contain multiple objects. Filtering with topk=1 is too strict and would lead to a small recall of the correct samples detection.

\paragraph{Data Augmentation}
For all data, we normalize the RGB-images by the mean: (0.485, 0.456, 0.406) and the standard variance (0.229, 0.224, 0.225). For training data, we perform random rotation of up to 10 degrees, randomly resize images to 224x224, apply random horizontal flip and random color jittering. This noise is needed in regular mean-teacher training. The jittering setting are: brightness=0.4, contrast=0.4, saturation=0.4, hue=0.1. The validation data are resized to 256x256 and randomly cropped to 224x224

\section{Losses}

\begin{figure}[t]
	\centering
	
	\begin{subfigure}{.30\textwidth}
		\begin{center}
			\includegraphics[width=1\linewidth]{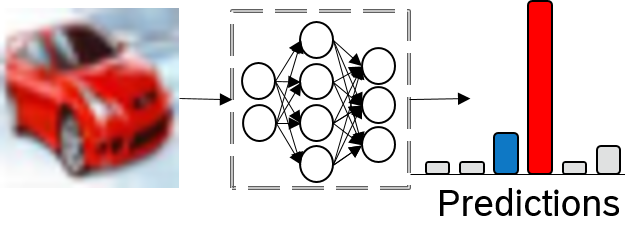}
			\caption{}
			\label{fig:noise-perc}
		\end{center}
	\end{subfigure}
	\begin{subfigure}{.30\textwidth}
		\begin{center}
			\includegraphics[width=1\linewidth]{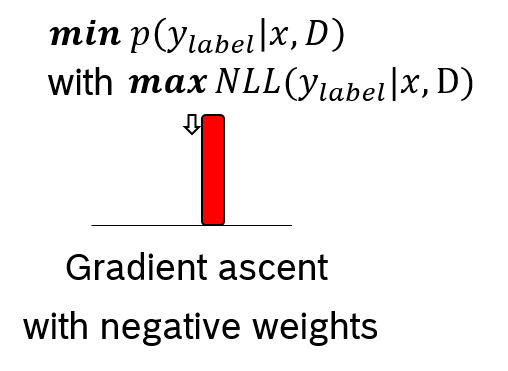}
			\caption{}          
			\label{fig:noise-recall}
		\end{center}
	\end{subfigure}
	\begin{subfigure}{.30\textwidth}
		\begin{center}
			\includegraphics[width=1\linewidth]{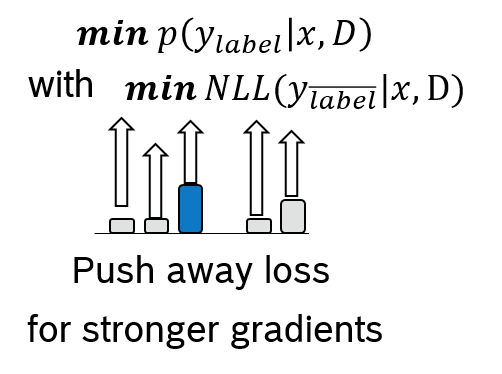}
			\caption{}          
			\label{fig:noise-recall}
		\end{center}
	\end{subfigure}            
	
	\caption{ Simple training losses to counter label noise. 
		(a) shows the prediction of a sample given a model. The red bar indicates the noisy label, blue the correct one. Arrows depict the magnitude of the gradients (b) Typical losses re-weighting schemes are not wrong but suffer from the gradient vanishing problem. Non-linear losses such as Negative-log-likelihood are not designed for gradient ascent. (c)Push-away-loss: as a simple baseline, we propose the intuitive push-away-loss to improve the gradients. We take this version as a strong baseline for a fair comparison.}
	\label{fig:lossess}
\end{figure}

For the learning of wrongly labeled samples, Fig. ~\ref{fig:lossess} shows the relationship between the typical reweighting scheme and our baseline push-away-loss. Typically, reweighting is applied directly to the losses with samples weights $w^{(i)}$ for each sample $i$ as shown in Eq. \ref{Eq:reweighting}

\begin{equation}
\min w^{(i)}_{j} NLL(y_{label}^{(i)}| x^{(i)},D) 
\label{Eq:reweighting}
\end{equation}
$D$ is the dataset,$x^{(i)}$ and  $y_{label}^{(i)}$  are the samples $i$ and its noisy label. $w^{(i)}_{j}$ is the samples weight for the sample $i$ at step $j$. Negative samples weights $w^{(i)}_{j}$ are often assigned to push the network away from the wrong labels. Let $w^{(i)}_{j}=-c^{(i)}_{j}$ with $c^{(i)}_{j}>0$, then we have: 
\begin{equation}
\min - c^{(i)}_{j} NLL(y_{label}^{(i)}| x^{(i)},D) 
\label{Eq:reweighting}
\end{equation}

Which results in:
\begin{equation}
\max c^{(i)}_{j} NLL(y_{label}^{(i)}| x^{(i)},D)
\label{Eq:reweighting}
\end{equation}
In other words, we perform gradient ascent for wrongly labeled samples. However, the Negative-log-likelihood is not designed for gradient ascent. Hence the gradients of wrongly labeled samples vanish if the prediction is too close to the noisy label. This effect is similar to the training of Generative Adversarial Network (GAN) \cite{goodfellow_generative_nodate}. In the GAN-framework, the generator loss is not simply set to the negated version of the discriminator's loss for the same reason.

Therefore, to provide a fair comparison with our framework, we suggest the push-away-loss $L_{Push-away}(y_{label}^{(i)},x^{(i)},D)$ with improved gradients as follows:

\begin{equation}
\min \frac{1}{|Y|-1}\sum_{y, y\neq y_{label}^{(i)}}  c^{(i)}_{j} NLL(y| x^{(i)},D)
\label{Eq:reweighting}
\end{equation}
Whereby $Y$ is the set of all classes in the training set. This loss has improved gradients to push the model away from the potentially wrong labels.
\end{document}